\documentclass[acmsmall]{acmart}
\AtBeginDocument{%
  \providecommand\BibTeX{{%
    \normalfont B\kern-0.5em{\scshape i\kern-0.25em b}\kern-0.8em\TeX}}}

\usepackage{hyperref}       
\usepackage{url}            
\usepackage{booktabs}       
\usepackage{amsfonts}       
\usepackage{nicefrac}       
\usepackage{microtype}      
\usepackage{xcolor}         

\usepackage{amsmath}
\usepackage{algorithm}
\usepackage{algpseudocode}
\usepackage{tabularx}
\usepackage{enumitem}
\usepackage{multirow}
\usepackage{subcaption}
\usepackage{graphicx}
\usepackage{wrapfig}

\usepackage{siunitx}
\newcolumntype{d}{S[table-format=3.2(3), separate-uncertainty]}

\newtheorem{definition}{Definition}

\newtheorem{assumption}{Assumption}
\newtheorem{problem}{Problem}
\makeatletter
\def\@opargbegintheorem#1#2#3{\trivlist
   \item[]{\bfseries #1\ #2\ (#3)} \itshape}
\makeatother

\makeatletter
\newcommand{\multiline}[1]{%
  \begin{tabularx}{\dimexpr\linewidth-\ALG@thistlm}[t]{@{}X@{}}
    #1
  \end{tabularx}
}
\makeatother 

\newcommand{\sysname}{SODIUM}

\begin{document}

\title{Towards Effective Semantic OOD Detection in Unseen Domains: A Domain Generalization Perspective}

\author{Haoliang Wang}
\email{haoliang.wang@utdallas.edu}
\affiliation{%
  \institution{The University of Texas at Dallas}
  \streetaddress{800 W. Campbell Road}
  \city{Richardson}
  \state{Texas}
  \country{USA}
  \postcode{75080-3021}
}

\author{Chen Zhao}
\email{chen\_zhao@baylor.edu}
\affiliation{%
  \institution{Baylor University}
  \streetaddress{One Bear Place \#97356}
  \city{Waco}
  \state{Texas}
  \country{USA}
  \postcode{76798-7356}
}

\author{Yunhui Guo}
\affiliation{%
  \institution{The University of Texas at Dallas}
  \streetaddress{800 W. Campbell Road}
  \city{Richardson}
  \state{Texas}
  \country{USA}
  \postcode{75080-3021}
}

\author{Kai Jiang}
\affiliation{%
  \institution{The University of Texas at Dallas}
  \streetaddress{800 W. Campbell Road}
  \city{Richardson}
  \state{Texas}
  \country{USA}
  \postcode{75080-3021}
}

\author{Feng Chen}
\affiliation{%
  \institution{The University of Texas at Dallas}
  \streetaddress{800 W. Campbell Road}
  \city{Richardson}
  \state{Texas}
  \country{USA}
  \postcode{75080-3021}
}

\renewcommand{\shortauthors}{Wang, et al.}

\begin{abstract}
  Two prevalent types of distributional shifts in machine learning are the covariate shift (as observed across different domains) and the semantic shift (as seen across different classes). Traditional OOD detection techniques typically address only one of these shifts. However, real-world testing environments often present a combination of both covariate and semantic shifts. In this study, we introduce a novel problem—semantic OOD detection across domains—which simultaneously addresses both distributional shifts. To this end, we introduce two regularization strategies: domain generalization regularization, which ensures semantic invariance across domains to counteract the covariate shift, and OOD detection regularization, designed to enhance OOD detection capabilities against the semantic shift through energy bounding. Through rigorous testing on three standard domain generalization benchmarks, our proposed framework, \sysname{}, showcases its superiority over conventional domain generalization approaches in terms of OOD detection performance. Moreover, it holds its ground by maintaining comparable InD classification accuracy.
\end{abstract}

\maketitle

\section{Introduction}
\begin{wrapfigure}{r}{8cm}
    \centering
    \includegraphics[width=0.95\linewidth]{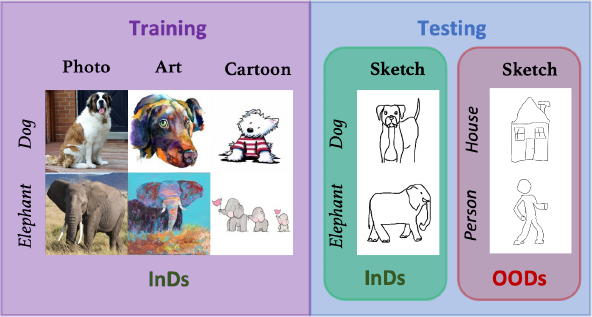}
    \caption{Illustration of the problem of semantic OOD detection across domains using the real-world \textsc{PACS} dataset \cite{li2017deeper}. Images labeled with ``Dog" and ``Elephant" are observed in training domains, Photo, Art, and Cartoon. The goal of the problem is to accurately detect semantic OODs (``House" and ``Person") in unseen test domains (Sketch).}
    \label{fig:problem}
\end{wrapfigure}

    Out-of-distribution (OOD) detection is a crucial aspect of machine learning techniques in many real-world applications, such as autonomous driving \cite{lee2017training}, cybersecurity \cite{grosse2017statistical}, fraud detection \cite{ruff2018deep}, and so on. In OOD detection problems, test data are mixed with data drawn \textit{i.i.d} from the same distribution as the training data, known as in-distribution (InD), and ones from out-of-distributions. 
The distribution shifts can be mainly sorted into covariate shifts, defined as OOD instances (OODs) sampled from distinct data domains where each is associated with a particular variation, and semantic shifts, wherein OODs are drawn from different classes. However, how to detect OODs with novel classes under distribution shifts in data variation remains unknown. As described in Figure \ref{fig:problem}, although images of ``\textit{Dog}" and ``\textit{Elephant}" in the sketch domain are new in data variation, they are known classes to the training data. The sketch ``\textit{House}" and ``\textit{Person}" are the true OODs that need to be detected, as they are semantically novel to InD instances (InDs).

The detection of semantic distribution shift due to the occurrence of new classes is the focal point of OOD detection tasks. The mainstream research on it can be divided into three areas: semantic anomaly detection (SAD) \cite{ruff2021unifying}, (one or) multi-class novelty detection (MND) \cite{sabokrou2018adversarially,bhattacharjee2020multi}, and open-set recognition (OSR) \cite{geng2020recent,zhao2023open}. Although SAD and MND aim to detect any test samples that do not fall into training classes, MND is supposed to be fully unsupervised, but SAD may have some abnormal training samples. 
OSR aligns well with the semantic OOD detection framework wherein it requires the multi-class predictor to simultaneously detect test samples if they are unknown classes or else to classify test samples if they are known classes to the training set. 
Despite existing endeavors that have achieved unprecedented success in semantic OOD detection, most of them assume that no covariate shift takes place during inference, meaning that data variations in test and training domains are unchanged. 

In this paper, we introduce a novel problem, namely \textit{semantic OOD detection under covariate shifts on data variations}. In the problem setting, training data consisting of finite domains, where each has a specific variation, are observed. The goal of it is to seek a good predictor where its outputs can be further used to identify semantic OODs in unseen test domains. However, training such a predictor is challenging:
\begin{itemize}
    \item The predictor is required to be domain-invariant and
    \item The outputs of it can be used for semantic OOD detection in unseen test domains.
\end{itemize}
To tackle the aforementioned challenges, technically, we propose a simple but effective approach, namely \sysname{}. Our method consists of three significant components: data disentanglement with semantic and variation factors, invariance of semantic representations, and generalization of pseudo-OODs by inter-class semantic mixup of InDs. 

Firstly, inspired by \cite{robey2021model,huang2018multimodal,zhao2023towards,zhao2021fairness,zhao2022adaptive}, we claim that inter-domain variation is solely characterized by the covariate shift due to $G$. We assume data over all domains are generated from an underlying transformation model $G$, which is decomposed of semantic and variation encoders and a decoder. The transformation model not only disentangles data into latent semantic and variation factors but, importantly, generates new data within synthetic domains with unchanged semantics but a different variation randomly sampled from its prior. With pre-trained $G$, a metric $R_{DG}$ is proposed with respect to semantic invariance. Specifically, it controls the distance in a representation space mapped from the featurizer of the predictor between an instance and its corresponding pairs in synthetic domains. We claim that, with $R_{DG}$, the featurizer captures semantic features accurately, and hence $R_{DG}$ is more practical for OOD detection due to semantic shift. Furthermore, since test domains containing OODs are inaccessible during training, to ensure the outputs of a predictor can be used for semantic OOD detection, pseudo-OODs are generated by mixing up InDs sampled from different classes in the latent semantic space. Lastly, another regularization term $R_{OOD}$ designed based on the energy scoring function \cite{liu2020energy} is proposed for semantic OOD detection. $R_{OOD}$ regulates the predictor by scoring instances between InDs in the training set and pseudo-OODs.

Our contributions are summarized:
\begin{itemize}
    \item In this paper, we proposed a novel problem that aims to detect semantic OOD across domains under covariate shifts on data variations. Moreover, we assume training data with multiple domains are observed, but test domains mixed with InDs and OODs are unseen.
    \item To tackle the problem, we introduce a simple but effective framework, namely \sysname{}. It consists of three main components: a transformation model and two regularizers added to the predictor.
    \item We empirically show that \sysname{} significantly outperforms baseline methods (various combinations between domain generalization backbones with OOD detectors) on three image benchmarks, including \textsc{ColoredMNIST}, \textsc{PACS}, and \textsc{VLCS}.
\end{itemize}

\section{Related Work}

\textbf{Semantic Out-of-Distribution Detection.}
The primary objective of semantic out-of-distribution detection tasks is to identify data that falls outside the semantic distribution of the training set. These tasks can go by various names depending on the context, including semantic anomaly detection (SAD), (one or) multi-class novelty detection (MND), and open set recognition (OSR), among others \cite{yang2021generalized}. However, at their core, they all share a common purpose: detecting semantic shifts between training and test sets. Within the realm of deep learning, this semantic shift is typically addressed in one of two spaces: the output space or the feature space.

Output-based methods, such as MSP \cite{hendrycks2016baseline}, ODIN \cite{liang2017enhancing}, and Energy \cite{liu2020energy}, tackle this issue by manipulating the outputs of the last layer. They operate under the presumption that inputs from out-of-distribution (OOD) sources tend to yield uniformly distributed class probabilities. In contrast, feature-based methods, like the Mahalanobis distance score \cite{lee2018simple}, Gram matrix-based score \cite{shama2019detecting}, and DDU \cite{mukhoti2021deep}, leverage feature representations from intermediate layers. This allows them to tap into a wealth of information to better distinguish OOD inputs. It's worth noting, however, that both output-based and feature-based methods assume that the training and test sets originate from the same domain, which is often violated in real-world scenarios.






\textbf{Domain Generalization} (DG) has gained significant traction in the machine learning community due to its promise of developing models with the capability to generalize across unseen domains. Broadly, the aim of DG is to train models on a set of source domains such that they perform well on a previously unseen target domain, without any form of adaptation. This is particularly crucial when the target domain data is not available during training.

Learning the semantic representations is also key to the success of some DG models. MBDG \cite{robey2021model} aims to capture domain-invariant features for robust generalization through penalizing deviated class distribution in the output space. While DDU \cite{zhang2022towards} explored the potential of disentangled representations to enhance the domain generalization capacity. Several other notable works include methodologies that emphasize domain-invariance by minimizing domain-specific information \cite{li2018learning}, utilizing adversarial techniques to reduce domain discrepancies \cite{ganin2016domain}, and learning domain-agnostic representations through meta-learning \cite{balaji2018metareg}. Yet, conventional domain generalization approaches typically overlook OOD inputs. Directly incorporating standard OOD detection techniques into domain generalization often yields suboptimal results.

\section{Preliminaries}

\textbf{Notations.} For clear interpretation, we list the notations used in this paper and their corresponding explanation shown in Table \ref{tab:notations}.
\begin{table}[!ht]
    \centering
    \caption{Important notations and corresponding descriptions.}
    \begin{tabular}{c|l}
       \toprule
        \textbf{Notations} & \textbf{Descriptions} \\
        \midrule
        $\mathcal{D}$ & entire dataset \\
        $\mathcal{E}, E, e$ & set of domain labels, number of all domains, index of domains \\
        $\mathcal{E}_{tr},\mathcal{E}_{te}$ & set of training and testing domains\\
        $\mathcal{D}_{tr}, \mathcal{D}_{te}$ & training and testing datasets\\
        $E_{tr}, E_{te}$ & numbers of training and testing domains\\
        $\mathcal{D}^e, |\mathcal{D}^e|$ & data of domain $e$, size of data of domain $e$  \\
        $\mathbf{x},y$ &  data instance and its label\\
        $\mathcal{X}$ & feature space \\
        $\mathcal{Y}_{tr}, \mathcal{Y}_{te}$ & label spaces of training and testing datasets \\
        $K$ & total number of training classes \\
        $y_{InD}, y_{OOD}$ & labels of testing data \\
        $f$ & the predictor \\
        $g,h$ & featurizer and classifier of the predictor $f$\\
        $\omega$ & OOD detector \\
        $\boldsymbol{\theta}, \boldsymbol{\theta}_g, \boldsymbol{\theta}_h$ & parameters of the predictor, the fraturizer and the classifier \\
        $\Theta$ & parameter space\\
        $G$ & the transformation model\\
        $E_s, E_v, D$ & semantic encoder, the variation encoder, and the decoder\\
        $\mathcal{S, V}$ & latent semantic and variation spaces\\
        $\mathbf{s}, \mathbf{v}$ & semantic and variation factors\\
        $R_{DG}$ & the domain generalization regularizer\\
        $d$ & distance metrics\\
        $R_{OOD}$ & the OOD regularizer\\
        $\lambda_1,\lambda_2$ & mixing parameters\\
        $\hat{\mathbf{s}}$ & semantic factors after mixing up\\
        $\hat{\mathbf{x}}$ & generated pseudo-OOD instance\\
        $\mathcal{D}_{pseudo}$ & dataset of pseudo-OODs\\
        $\xi$ & empirical threshold qualifying pseudo-OODs\\
        $\mathbf{x}_{in},y_{InD}$ & InD instances in the training dataset\\
        $\mathbf{x}_{out},y_{OOD}$ & pseudo-OOD instances\\
        $m_{in}, m_{out}$ & margins regularizing InD and pseudo-OOD energy scores\\
        $E_g$ & energy function\\
        $T$ & temperature parameter\\
        $\beta_1,\beta_2$ & Lagrangian multipliers (dual variables)\\
        $\gamma_1,\gamma_2$ & margins of the two regularizers\\
        $\eta_p,\eta_{dg},\eta_{ood}$ & learning rates of primal and dual parameters\\
       \bottomrule
    \end{tabular}
    \label{tab:notations}
\end{table}

\textbf{Problem Setting.} Given a dataset $\mathcal{D}$, we consider a set of domains $\mathcal{E}=\{e\}_{e=1}^E\in\mathbb{N}$, where each corresponds to a data subset $\mathcal{D}^{e}=\{(\mathbf{x}^{e}_i,y^{e}_i)\}_{i=1}^{|\mathcal{D}^{e}|}$ with a specific data variation and $\mathcal{D}=\{\mathcal{D}^{e}\}_{e=1}^E$. 
The dataset $\mathcal{D}$ is partitioned into multiple training domains $\mathcal{E}_{tr}\in\mathcal{E}$ and test domains $\mathcal{E}_{te}=\mathcal{E}\backslash\mathcal{E}_{tr}$. Training domains are associated with data $\mathcal{D}_{tr}=\{\mathcal{D}^e\}_{e=1}^{E_{tr}}\in(\mathcal{X}\times\mathcal{Y}_{tr})$, where we
denote $\mathcal{X}\in\mathbb{R}^d$ as the feature space and $\mathcal{Y}_{tr}=\{1,\cdots,K\}$ as the label space. $K$ is the total number of observed classes in $\mathcal{D}_{tr}$. Test domains $\mathcal{E}_{te}$ are associated with data $\mathcal{D}_{te}=\mathcal{D}\backslash\mathcal{D}_{tr}\in(\mathcal{X}\times\mathcal{Y}_{te})$, where $\mathcal{Y}_{te}=\{y_{OOD}=0, y_{InD}=1\}$.

Generally speaking, the aim of OOD detection problems \cite{yang2021generalized} is to train a predictor using $\mathcal{D}_{tr}$, such that, for any datapoint in $\mathcal{D}_{te}$, if it is an observation from $\mathcal{D}_{tr}$, the predictor detects it as InD, OOD otherwise. Without loss of generality, let all detected OODs during inference be allocated to the $y_{OOD}$ class and InDs to $y_{InD}$ that collapses $K$ classes from $\mathcal{Y}_{tr}$ to 1.

\textbf{Problem Formulation.} As shown in Figure \ref{fig:problem}, since each domain has a specific data variation, in this paper, we propose a novel problem, namely \textit{semantic OOD detection across domains}, wherein the goal is to learn a predictor $f:\mathcal{X}\times\Theta\rightarrow\mathcal{Y}_{tr}$ on \textit{observed} training domains $\mathcal{E}_{tr}$. The predictor $f$ can be further generalized on \textit{unseen} test domains $\mathcal{E}_{te}$ to identify OODs semantically through a given detector $\omega:\mathbb{R}^q\rightarrow\mathcal{Y}_{te}$. 
\begin{problem}[Semantic OOD Detection across Domains]
\label{prob:Problem}
    Let $\mathcal{E}_{tr}\subset\mathcal{E}$ be a finite subset of data domains and assume that we have access to its corresponding data $\mathcal{D}_{tr}=\{\mathcal{D}^{e}\}_{e=1}^{E_{tr}}$ where, for each $e\in\mathcal{E}_{tr}$, $\mathcal{D}^{e}=\{(\mathbf{x}^{e}_i,y^{e}_i)\}_{i=1}^{|\mathcal{D}_{e}|}$ and each has distinct data variations. 
    Given a loss function $\mathcal{L}_{CE}:\mathcal{Y}\times\mathcal{Y}\rightarrow\mathbb{R}$, the goal is to learn a predictor $f:\mathcal{X}\times\Theta\rightarrow\mathcal{Y}_{tr}$ that minimizes the worst-case risk over observed training domains
    \begin{align}
        \label{eq:problem}
        \min_{\boldsymbol{\theta}\in\Theta} \: \max_{e\in\mathcal{E}_{tr}} \:\mathbb{E}_{(\mathbf{x}^e,y^e)\sim\mathcal{D}_{tr}} \mathcal{L}_{CE} ( f(\mathbf{x}^e,\boldsymbol{\theta}),y^e )
    \end{align}
    where $\mathcal{L}_{CE}:\mathcal{Y}_{tr}\times\mathcal{Y}_{tr}\rightarrow\mathbb{R}$ denotes the cross-entropy loss.
    Let $\mathcal{E}_{te}=\mathcal{E}\backslash\mathcal{E}_{tr}$ be the unseen test domains and assume that they are inaccessible to $\mathcal{D}^{te}=\{\mathcal{D}^e\}_{e=1}^{E_{te}}$ at training. During inference, given an OOD detector $\omega:\mathbb{R}^q\rightarrow\mathcal{Y}_{te}$ and the learned predictor $f$, the OOD detection risk $\mathbb{E}_{(\mathbf{x}^e,y^e)\sim\mathcal{D}_{te}} \omega(f(\mathbf{x}^e),y^e)$ with respect to data class is minimized.
\end{problem}
The key of Problem~\ref{prob:Problem} is to seek a predictor $f$ where its outputs can be used for identifying novel classes in shifted domains with a given semantic OOD detector $\omega$. To this end, a good $f$ must meet the following two requirements. 
\begin{enumerate}
    \item $f$ is invariant and generalizable across domains. 
    \item The output of $f$ can be further used by $\omega$ for semantic OOD detection. Since we do not assume data from $\mathcal{E}_{te}$ is accessible, and data with novel classes only lie in $\mathcal{D}_{te}$, it makes Problem~\ref{prob:Problem} challenging to solve.
\end{enumerate}

\textbf{Remarks.} In general, generalized OOD detection tasks are characterized by the distribution shift across domains, which may be categorized by various types of shifts \cite{robey2021model,gui2022good}. In this paper, we restrict the scope that inter-domain variation is due solely to covariate shift \cite{robey2021model}, that domain variation is due to differences between the set of marginal distributions over $\{\mathbb{P}(X^e)\}_{e\in\mathcal{E}}$, through an underlying transformation model $G:\mathcal{X}\times\mathcal{E}\rightarrow\mathcal{X}$. Our goal is to detect OODs in testing domains due to semantic shifts.

Moreover, semantic OOD detection often fails by providing high-confidence predictions while being woefully incorrect, especially when training data contains spurious correlations between classes and domain variations \cite{ming2022impact}. These high-confidence predictions are frequently produced by softmaxs because softmax probabilities are computed with the fast-growing exponential function \cite{hendrycks2016baseline}. Thus, minor additions to the softmax input can substantially change the output distribution. 

In deep learning, a predictor $f:=g\circ h$ is decomposed of a featurizer $g:\mathcal{X}\times\Theta\rightarrow\mathbb{R}^r$ and a classifier $h:\mathbb{R}^r\times\Theta\rightarrow\mathbb{R}^K$. Therefore, we assume the instance-conditional distributions of data features through $g$ are stable across domains.
\begin{assumption}[Invariance across Domains]
\label{assump:invariance}
    Denote $g:\mathcal{X}\times\Theta\rightarrow\mathbb{R}^r$ and $h:\mathbb{R}^r\times\Theta\rightarrow\mathbb{R}^K$ as the featurizer and the classifier of a predictor $f$, respectively. We assume that inter-domain variation is solely characterized by the shift in the marginal distributions over $\{\mathbb{P}(X^e)\}_{e\in\mathcal{E}}$. As a consequence, we assume that the distributions of data features via $g$ are stable across domains. Formally, for any domain $e_1,e_2\in\mathcal{E}$ and $y\in\mathcal{Y}_{tr}$, it holds that
    \begin{align*}
        \mathbb{P}(g(X^{e_1},\boldsymbol{\theta}_g)|X^{e_1}=\mathbf{x}^{e_1},Y^{e_1}=y) = 
        \mathbb{P}(g(X^{e_2},\boldsymbol{\theta}_g)|X^{e_2}=G(\mathbf{x}^{e_1}, e_2), Y^{e_2}=y)
    \end{align*}
\end{assumption}
Under Assumption~\ref{assump:invariance}, we assume that InDs for each domain $e\in\mathcal{E}$ is generated from an underlying transformation model $G$. We hence introduce a definition of semantic invariance with respect to the variation captured by $G$.
\begin{definition}[Semantic $G$-invariance]
\label{def:semantic_g_inv}
    Given $G$, a predictor $f=g\circ h$ is semantic invariance if it holds $g(\mathbf{x}^{e_1},\boldsymbol{\theta}_g)=g(\mathbf{x}^{e_2},\boldsymbol{\theta}_g)$ almost surely when $\mathbf{x}^{e_2}=G(\mathbf{x}^{e_1},e_2),\mathbf{x}^{e_1}\sim\mathbb{P}(X^{e_1}), \mathbf{x}^{e_2}\sim\mathbb{P}(X^{e_2})$ and $\forall e_1,e_2\in\mathcal{E}$.
\end{definition}
This definition is designed to enforce invariance directly on the output of the featureizer $g$ made by $f$. 

Moreover, existing domain generalization endeavors \cite{robey2021model,vapnik1999nature} encompass the semantic $G$-invariance and propose state-of-the-art solutions by mainly focusing on the disentanglement of the variation of data across domains into latent spaces. We, therefore, make the following assumption.
\begin{assumption}[Multiple Latent Factors]
\label{assump:mlfactors}
    Given an InD $(\mathbf{x}^e, y^e)$ from a particular domain $e\in\mathcal{E}$, we assume it is generated from two factors
    \begin{itemize}
        \item a latent semantic factor $\mathbf{s}\in\mathcal{S}$, where $\mathcal{S}=\{\mathbf{s}_{y^e=1},\cdots, \mathbf{s}_{y^e=K}\}$;
        \item a latent variation factor $\mathbf{v}^e\in\mathcal{V}$, where $\mathbf{v}^e$ is specific to the individual domain $e$.
    \end{itemize}  
\end{assumption}
In the following section, we propose two regularization terms added on the predictor $f$, $R_{DG}$ and $R_{OOD}$, where each corresponds to an aforementioned requirement, respectively.

\section{Methodology}

\begin{figure*}[t]
    \centering
    \includegraphics[width=\linewidth]{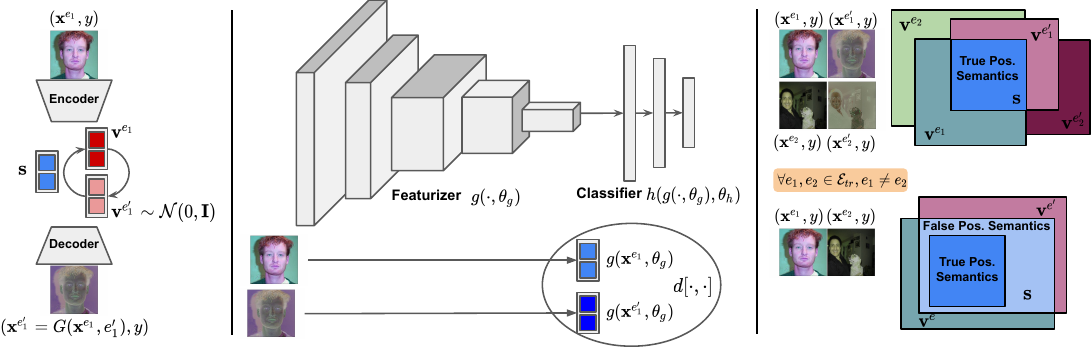}
    \caption{Illustration of Semantic $G$-invariance. (\textbf{Left}) An underlying transformation model $G$ is used to generate a new instance $(\mathbf{x}^{e'_1},y)$ from its original input $(\mathbf{x}^{e_1},y)$ by randomly sampling a variation factor $\mathbf{v}^{e'_1}$ from its prior $\mathcal{N}(0,\mathbf{I})$. (\textbf{Middle}) A predictor $f$ is decomposed of a featurizer $g$ and a classifier $h$. Under Assumption \ref{assump:invariance}, we propose a regularizer $R_{DG}$ that controls the distance of the outputs of $g$ by giving two instances within the same class. (\textbf{Right}) Without instances from synthetic domains, the featurizer may not capture the semantics of inputs accurately, due to finite training domains, where false positive semantics are mistakenly taken for downstream semantic OOD detection tasks. 
    }
    \label{fig:intuition}
\end{figure*}

\textbf{Data Disentanglement via $G$.} Under Assumptions~\ref{assump:invariance} and \ref{assump:mlfactors} and same to the network (MIITN) in \cite{robey2021model}, in practice, the underlying transformation model $G$ is designed with two goals and it consists of three components, a semantic encoder $E_s:\mathcal{X}\rightarrow\mathcal{S}$, a variation encoder $E_v:\mathcal{X}\rightarrow\mathcal{V}$, and a decoder $D:\mathcal{S}\times\mathcal{V}\rightarrow\mathcal{X}$. The first goal is to disentangle an input $(\mathbf{x}^e,y)$ into a semantic factor $\mathbf{s}=E_s(\mathbf{x}^e)$ and a variation factor $\mathbf{v}^e=E_v(\mathbf{x}^e)$. Such factors are further used to recover the input through $D(\mathbf{s},\mathbf{v}^e)$ in the particular domain $e$. Another goal of $G$ is to generate an instance $(\mathbf{x}^{e'},y)$ in a synthetic domain $e'$. Thus, given $\mathbf{s}$ and $\mathbf{v}^e$ at the output of disentangled factors of $(\mathbf{x}^e, y)$, we generate instances by only replacing $\mathbf{v}^e$ with a random sampled $\mathbf{v}^{e'}\sim \mathcal{N}(0,\mathbf{I})$ to produce $(\mathbf{x}^{e'},y)$.

\textbf{$G$-Invariance of Semantic Representations.} Under Assumption~\ref{assump:invariance}, we assume the distributions of data features via $g$ are stable across domains. To enforce the semantic invariance between the input $(\mathbf{x}^{e},y)$ and the generated $(\mathbf{x}^{e'},y)$ in domain $e$ and $e'$, respectively. A domain generalization regularization is proposed, denoted as $R_{DG}$, where
\begin{align}
\label{eq:reg_dg}
    R_{DG} = \mathbb{E}_{(\mathbf{x}^e, y)\sim\mathcal{D}_{tr}} d\Big[g(\mathbf{x}^e, \boldsymbol{\theta}_g), g\big(D(E_s(\mathbf{x}^e), \mathbf{v}^{e'}), \boldsymbol{\theta}_g\big) \Big]
\end{align}
where $\mathbf{v}^{e'}\sim\mathcal{N}(0,\mathbf{I})$. In our experiments, we use $\ell_2$-norm for the distance function $d:\mathbb{R}^r\times\mathbb{R}^r\rightarrow\mathbb{R}$ in $R_{DG}$.

\textbf{Remarks.} The regularizer $R_{DG}$ is proposed for two purposes. 
As shown in Figure \ref{fig:intuition}, (1) It helps $g$ capture semantic features more accurately. With limited training domains at hand, $R_{DG}$ aims to minimize the distance of outputs of the featurizer $g$ between data and their augmented pair in synthetic domains generated from $G$. 
Ideally, the more data generated from synthetic domains, the more accurately $g$ captures semantic features. 
(2) Due to the first purpose, $R_{DG}$ is more practical for OOD detection due to semantic shift.

\textbf{Generation of Pseudo-OODs by Inter-Class Semantic Mixup of InDs.} As stated in Problem~\ref{prob:Problem}, semantic OOD detection aims to identify instances with novel classes. However, such OOD instances are often unknown in advance. 
To this end, we generate pseudo-OOD instances via in-distribution inter-class mixup \cite{zhang2017mixup} within semantic space, where semantic factors of such pseudo-OOD instances are a convex combination of semantic factors of two instances with different class labels. Formally, given $(\mathbf{x}^{e_1},y_1)$ and $(\mathbf{x}^{e_2},y_2)$, where $\forall e_1,e_2\in\mathcal{E}_{tr}$, $y_1\neq y_2$ but we do not require $e_1\neq e_2$, inter-class mixup obtains a new augmented instance in the semantic space ${(\hat{\mathbf{x}}},y_{OOD})$ for training the predictor $f$ described in Eq.(\ref{eq:mixup}).
\begin{align}
\label{eq:mixup}
    \hat{\mathbf{s}} = \lambda_1\cdot \mathbf{s}_1 + \lambda_2\cdot \mathbf{s}_2 \quad \text{and} \quad \hat{\mathbf{x}} = D(\hat{\mathbf{s}}, \mathbf{v}^{e'})
\end{align}
where $\lambda_1, \lambda_2\in[-1,1]$ are mixing parameters. In the same fashion of inter-class mixup \cite{zhang2017mixup} $\lambda_1,\lambda_2\sim Beta(\alpha,\alpha)$ and for our experiments we set $\alpha=1$.
$\mathbf{s}_1=E_s(\mathbf{x}^{e_1})$ and $\mathbf{s}_2=E_s(\mathbf{x}^{e_2})$ are semantic factors for $\mathbf{x}^{e_1}$ and $\mathbf{x}^{e_2}$, respectively, encoded by the pre-trained semantic encoder $E_s$ in $G$. The pseudo-OOD instance is generated by the mixed semantic factor $\hat{\mathbf{s}}$ with randomly sampled $\mathbf{v}^{e'}$ through the pre-trained decoder $D$. 


\textbf{Screening Qualified Pseudo-OODs using GDA.} Indeed, generated pseudo-OODs through mixing up InDs in the latent space may not qualify for semantic OOD detection due to the choice of $\lambda_1$ and $\lambda_2$. To ensure the quality of pseudo-OODs, inspired by \cite{mukhoti2021deep}, we use Gaussian Discriminant Analysis (GDA) \cite{hastie1996discriminant}, a GMM $p(y,\mathbf{s})$ with a single Gaussian component per InD class, and fit each class component by computing the empirical mean and covariance, per class, of the semantic factors $\mathbf{s}=E_s(\mathbf{x})$, which are the outputs of the semantic encoder computed on samples over all training domains. 
Note that we do not require pseudo-OODs to fit these and, following \cite{mukhoti2021deep}, we use a separate covariance matrix for each class. Fitting a GDA on the latent semantic space, thus requires no further training and only requires a single forward pass through the training set. Furthermore, generated pseudo-OODs are evaluated by the fitted GDA. Qualified pseudo instances are added to $\mathcal{D}_{pseudo}$ if the density score $p(y,\mathbf{s}),\forall y\in\mathcal{Y}_{in}$ of each InD class is smaller than an empirical threshold $\xi>0$.  

\textbf{Energy Scoring based Semantic OOD Detection.} Since semantic OOD detection is a classification problem that relies on a score to differentiate between InD and OOD instances, a scoring function, such as the Energy function \cite{lecun2006tutorial}, is commonly used to produce distinguishable values between in- and out-of-distribution. Therefore, the semantic OOD detection regularization term $R_{OOD}$ is defined.
\begin{align}
\label{eq:reg_ood}
    R_{OOD} 
    = \mathbb{E}_{(\mathbf{x}_{in},y_{InD})\sim\mathcal{D}_{tr}}\big( \max(0,E_g(\mathbf{x}_{in},\boldsymbol{\theta})-m_{in}) \big)^2
    + \mathbb{E}_{(\mathbf{x}_{out},y_{OOD})\sim\mathcal{D}_{pseudo}}\big( \max(0,m_{out}-E_g(\mathbf{x}_{out},\boldsymbol{\theta})) \big)^2 \nonumber
\end{align}
where $\mathcal{D}_{pseudo}=\{\hat{\mathbf{x}}_i, y_{OOD}\}_{i=1}^m$ is the auxiliary pseudo-OOD data generated through semantic inter-domain mixup and examined by fitted GMM. $m_{in}$ and $m_{out}$ are margins regularizing InD and OOD energy scores.
$E_g:\mathcal{X}\times\Theta\rightarrow\mathbb{R}$ is the energy function introduced in \cite{liu2020energy}, defined as 
\begin{align*}
    E_g(\mathbf{x},\boldsymbol{\theta})=-T \cdot \log\sum\nolimits_{i=1}^K \exp^{f_i(\mathbf{x}, \boldsymbol{\theta})/T}
\end{align*}
where $T$ is the temperature parameter and $f_i(\mathbf{x}, \boldsymbol{\theta})$ indicates the logit corresponding to the $i$-th class label.

In essence, given a semantic OOD detector $\omega$, the total loss function is denoted
\begin{align}
    \ell(\boldsymbol{\theta}) 
    = \mathbb{E}_{(\mathbf{x}^e,y^e)\sim\mathcal{D}_{tr}}\mathcal{L}_{CE}(f(\mathbf{x}^e,\boldsymbol{\theta}), y^e) +\beta_1\cdot R_{DG} + \beta_2\cdot R_{OOD}
\end{align}
where $\beta_1,\beta_2>0$ are Lagrangian multipliers. $\gamma_1,\gamma_2>0$ are margins determined empirically.


\begin{algorithm}[t]
    \caption{\sysname{}}
    \label{alg:our_alg}
    \multiline{%
    \textbf{Require}: A pre-trained model $G=\{E_s,E_v, D\}$, fitted GDA using $\{(E_s(\mathbf{x}_i),y_i)\}_{i=1}^{|\mathcal{D}_{tr}|}$, and the energy score function $E_g$.\\
    }
    \textbf{Require}: Mixing parameters $\lambda_1,\lambda_2$, the threshold of GDA $\xi$, energy scoring margins $m_{in},m_{out}$, the temperature parameter $T$, regularization margins $\gamma_1,\gamma_2$, primal and dual learning rate $\eta_p,\eta_{dg},\eta_{ood}$
\begin{algorithmic}[1]
    \Repeat
        \For{minibatch $\{(\mathbf{x}_i,y_i)\}_{i=1}^m$ in training data $\mathcal{D}_{tr}$}
        \State $\mathcal{L}_{cls}\leftarrow(1/m)\sum_{i=1}^m\mathcal{L}_{CE}(f(\mathbf{x}_i,\boldsymbol{\theta}),y_i)$
        \State $(\Tilde{\mathbf{x}}_i,y_i)\leftarrow\textsc{DataAug}(\mathbf{x}_i,y_i),\: \forall i\in[m]$
        \State $R_{DG}\leftarrow(1/m)\sum_{i=1}^m d[g(\mathbf{x},\boldsymbol{\theta}_g), g(\Tilde{\mathbf{x}},\boldsymbol{\theta}_g)]$
        \State $\mathcal{D}_{pseudo}\leftarrow\emptyset$
        \For{each $(\mathbf{x}_i,y_i)$ in the minibatch}
            \State \multiline{%
                Sample a datapoint $(\mathbf{x}_j,y_j)$ from $\mathcal{D}_{tr}$ where $y_i\neq y_j$}
            \State $\hat{\mathbf{s}}=\lambda_1\cdot E_s(\mathbf{x}_i) + \lambda_2\cdot E_s(\mathbf{x}_j)$
            \If{GDA($\hat{\mathbf{s}})<\xi$}
                \State \multiline{%
                    Add $(D(\hat{\mathbf{s}},\mathbf{v}),y_{OOD})$ to $\mathcal{D}_{pseudo}$ where $\mathbf{v}\sim\mathcal{N}(0,\mathbf{I})$}
            \EndIf
        \EndFor
        \State $n\leftarrow|\mathcal{D}_{pseudo}|$
        \State \multiline{%
            $R_{OOD}\leftarrow(1/m)\sum_{i=1}^m( \max(0,E_g(\mathbf{x}_i,\boldsymbol{\theta})-m_{in}))^2+(1/n)\sum_{l=1}^n( \max(0,m_{out}-E_g(\mathbf{x}_l,\boldsymbol{\theta})) )^2$
        }
        \State $\ell(\boldsymbol{\theta})\leftarrow\mathcal{L}_{cls}+\beta_1\cdot R_{DG} + \beta_2\cdot R_{OOD}$
        \State $\boldsymbol{\theta}\leftarrow\text{Adam}(\ell(\boldsymbol{\theta}),\boldsymbol{\theta},\eta_p)$
        \State $\beta_1\leftarrow\max\{0, \beta_1+\eta_{dg}\cdot(R_{DG}-\gamma_1)\}$
        \State $\beta_2\leftarrow\max\{0, \beta_2+\eta_{ood}\cdot(R_{OOD}-\gamma_2)\}$
        \EndFor
    \Until{$\text{convergence}$}

    \Procedure{\textsc{DataAug}}{$\mathbf{x},y$}
        \State $\mathbf{s}\leftarrow E_s(\mathbf{x})$, $\mathbf{v}\leftarrow E_v(\mathbf{x})$
        \State Sample $\mathbf{v}'\sim\mathcal{N}(0,\mathbf{I})$
        \State \textbf{return } $(D(\mathbf{s},\mathbf{v}'),y)$
    \EndProcedure
\end{algorithmic}
\end{algorithm}

\textbf{\sysname{}: A Practical Approach.} In previous sections, to learn a good predictor $f$ that can generalize across domains and further be used for semantic OOD detection in unseen testing set by a given detector, we propose two regularization terms added on the cross-entropy loss during training. Eq.(\ref{eq:problem}) is thus reformulated to 
\begin{align}
\label{eq:reformulation}
    \min_{\boldsymbol{\theta}\in\Theta} \mathbb{E}_{(\mathbf{x}^e,y^e)\sim\mathbf{D}_{tr}} \ell(\boldsymbol{\theta})
\end{align}
Notice that Eq.(\ref{eq:reformulation}) is not a composite optimization problem, meaning that the inner maximization has been removed from Eq.(\ref{eq:problem}). To optimize, an effective approach is proposed in Algorithm \ref{alg:our_alg}.


In lines 22-26, we describe the \textsc{DataAug} procedure that takes a datapoint $(\mathbf{x},y)$ as input and returns an augmented example $(\mathbf{x}',y)$ from a synthetic domain as output, where $\mathbf{x}'=D(\mathbf{s},\mathbf{v}')$ and its variation factor $\mathbf{v}'$ randomly sampled from $\mathcal{N}(0,\mathbf{I})$ encodes a new synthetic domain. 
Lines 2-20 demonstrate the main training loop for \sysname{}. 
The domain generalization regularization $R_{DG}$ is computed in line 5. 
In lines 8-12, qualified pseudo-OODs are generated using inter-class InDs by mixing up their semantic factors. 
Energy score-based regularizer $R_{OOD}$ is shown in line 15. The primal parameter $\boldsymbol{\theta}$ and the dual parameters $\beta_1,\beta_2$ are updated in lines 17-19.

\section{Experiments}
    \textbf{Datasets.} We evaluated \sysname{} on three standard domain generalization benchmarks: \textsc{ColoredMNIST} \cite{arjovsky2019invariant} (3 domains, 70,000 samples, 2 classes), \textsc{PACS} \cite{li2017deeper} (4 domains, 9,991 samples, 7 classes), and \textsc{VLCS} \cite{torralba2011unbiased} (4 domains, 10,729 samples, 5 classes).

\textsc{ColoredMNIST} is a variant of the popular MNIST dataset of handwritten digits. It introduces color as a spurious feature, thereby creating three distinct domains for evaluating domain generalization capabilities. The three domains: [+90\%, +80\%, -90\%] are characterized by different levels of digit color and label correlations. The original 10 digits are split into two classes: digits from 0-4 are categorized with label 0, while digits from 5-9 receive label 1. Note that there is a 25\% manually injected error in the binary labels, which makes \textsc{ColoredMNIST} challenging for the domain generalization task.

\textsc{PACS} dataset comprises images sourced from four discrete domains: Art Painting (A), Cartoon (C), Photo (P), and Sketch (S). Each domain offers a wide array of images showcasing diverse objects, scenes, and individuals. These images are categorized into seven different categories, namely dog, elephant, giraffe, guitar, horse, house, and person.

\textsc{VLCS} is a collection of four well-established datasets
commonly used in computer vision research: Caltech101 (C), LabelMe (L), SUN09 (S), and VOC2007 (V). These datasets, when brought together, offer an extensive variety of images across diverse
contexts and subjects, making them an ideal test bed for assessing
domain generalization capabilities. These images are categorized into five classes: person, dog, chair, car, and bird.

\begin{figure}
\centering
\begin{subfigure}{.3\textwidth}
  \centering
  \includegraphics[width=.9\linewidth]{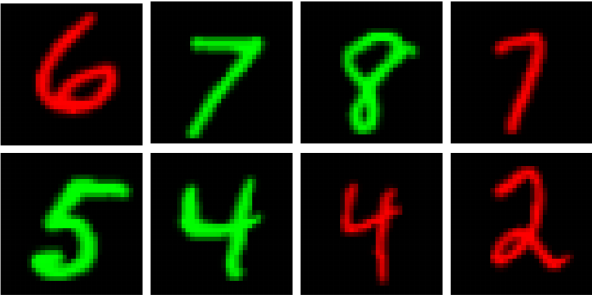}
  \caption{\textsc{ColoredMNIST}}
  \label{fig:coloredmnist}
\end{subfigure}%
\begin{subfigure}{.34\textwidth}
  \centering
  \includegraphics[width=.9\linewidth]{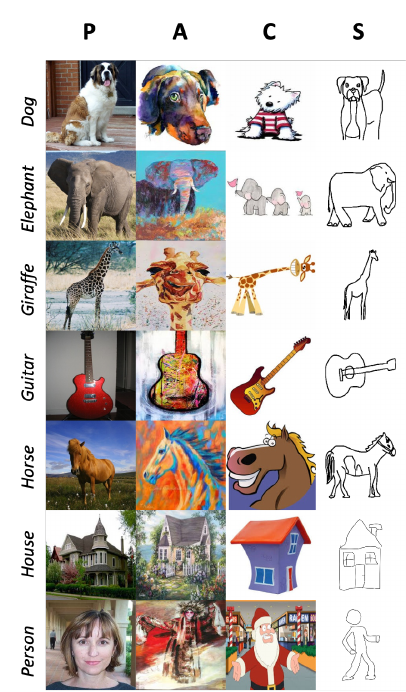}
  \caption{\textsc{PACS}}
  \label{fig:pacs}
\end{subfigure}
\begin{subfigure}{.34\textwidth}
  \centering
  \includegraphics[width=1\linewidth]{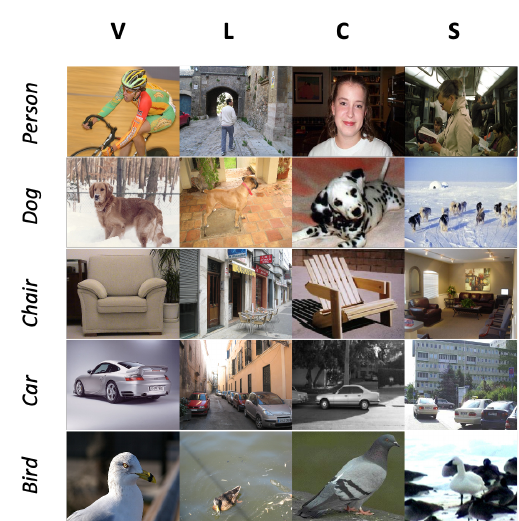}
  \caption{\textsc{VLCS}}
  \label{fig:vlcs}
\end{subfigure}
\caption{Samples from each dataset. (a) \textsc{ColoredMNIST} domains are different levels of digit color and label correlation: +90\%, +80\%, and -90\%; (b) \textsc{PACS} domains are: Photo (P), Art Painting (A), Cartoon (C), and Sketch (S); (c) \textsc{VLCS} domains are: VOC2007 (V), LabelMe (L), Caltech101 (C), and SUN09 (S).}
\label{fig:data-example}
\end{figure}

Some examples of \textsc{ColoredMNIST}, \textsc{PACS}, and \textsc{VLCS} are shown in Figure \ref{fig:data-example}.

\textbf{Environment and Experiment Setup.}
The environment we used: Python: 3.9.13, PyTorch: 1.12.1, Torchvision: 0.13.1, CUDA: 11.6, CUDNN: 8302, NumPy: 1.23.1, PIL: 9.2.0.
Experiment setup: Our experimental framework, tailored for semantic OOD detection across domains, deviates from the conventional setups for OOD detection and domain generalization. Uniquely, in our configuration, the OOD samples are always sourced from an unseen test domain. For each dataset, we designated one class as OOD and iterate our experiments with different randomly chosen OOD classes until a minimum of 40\% of classes have been designated as OOD. To further mitigate the impact of randomness, each set of experiments is conducted in two separate trials with different seeds. The results reported in the paper represent mean and standard error across these trials and all OOD class choices. Furthermore, given the multiple OOD detection methods evaluated in this study, and in an effort to present the results in a concise manner, for the OOD detection performance we report the averaged values across all test domains unless specified otherwise.

These experiments were conducted using the DomainBed \cite{gulrajani2020search} framework to ensure consistent assessment. We conducted a random search across the hyperparameter distribution for each algorithm and test domain, executing 20 runs in the process. 

\textbf{Model Selection.} Commonly used model selection strategies include training-domain validation, leave-one-domain-out cross-validation, and test-domain validation. Training-domain validation holds out a portion of the training data as the validation set, but essentially, it assumes that the test domain adheres to the same distribution as the training domains, which contradicts the concept of domain generalization. Test-domain validation uses part of the test set for model selection, but in practice, we typically lack any information about the target domain. Leave-one-domain-out cross-validation utilizes a held-out training domain as a validation set. This strategy presumes that training and test domains follow a meta-distribution over domains, which more closely aligns with real-world scenarios.


\textbf{OOD Detectors and Baselines.} Regarding OOD detection methods, we selected four commonly used OOD detection algorithms: MSP \cite{hendrycks2016baseline}, Energy \cite{liu2020energy}, DDU \cite{mukhoti2021deep}, OCSVM \cite{scholkopf1999support}. Each of these exploits different features and heuristics. Specifically, MSP and Energy exploit the class probabilities by assuming a uniform distribution for OOD examples, on the other hand, DDU approaches this problem by estimating the density in the feature space, and the goal of OCSVM is to find the optimal separating hyperplane in a high-dimensional space. Using these OOD detection methods, we juxtaposed \sysname{}'s performance in terms of OOD detection and classification accuracy against four established domain generalization baselines: ERM \cite{vapnik1999nature}, IRM \cite{arjovsky2019invariant}, Mixup \cite{yan2020improve}, and MBDG \cite{robey2021model}.

\textbf{Evaluation Metrics.}
To evaluate domain generalization performance, we simply report the classification accuracy. Meanwhile, to evaluate OOD detection performance, we use AUROC (Area Under the Receiver Operating Characteristic Curve). AUROC is a commonly used metric for evaluating the effectiveness of binary classification models. It measures the model's capacity to differentiate between positive and negative classes by computing the area under the ROC curve. This value can range from 0 to 1, with a perfect model having an AUROC of 1 and a random model having an AUROC of 0.5.

Note that the results reported in the paper are the mean and standard error across trials and all OOD class choices. And for the OOD detection performance evaluation, the AUROC values are further averaged across all test domain selections to save space.

\begin{figure}[!t]
    \centering
    \includegraphics[width=0.8\linewidth]{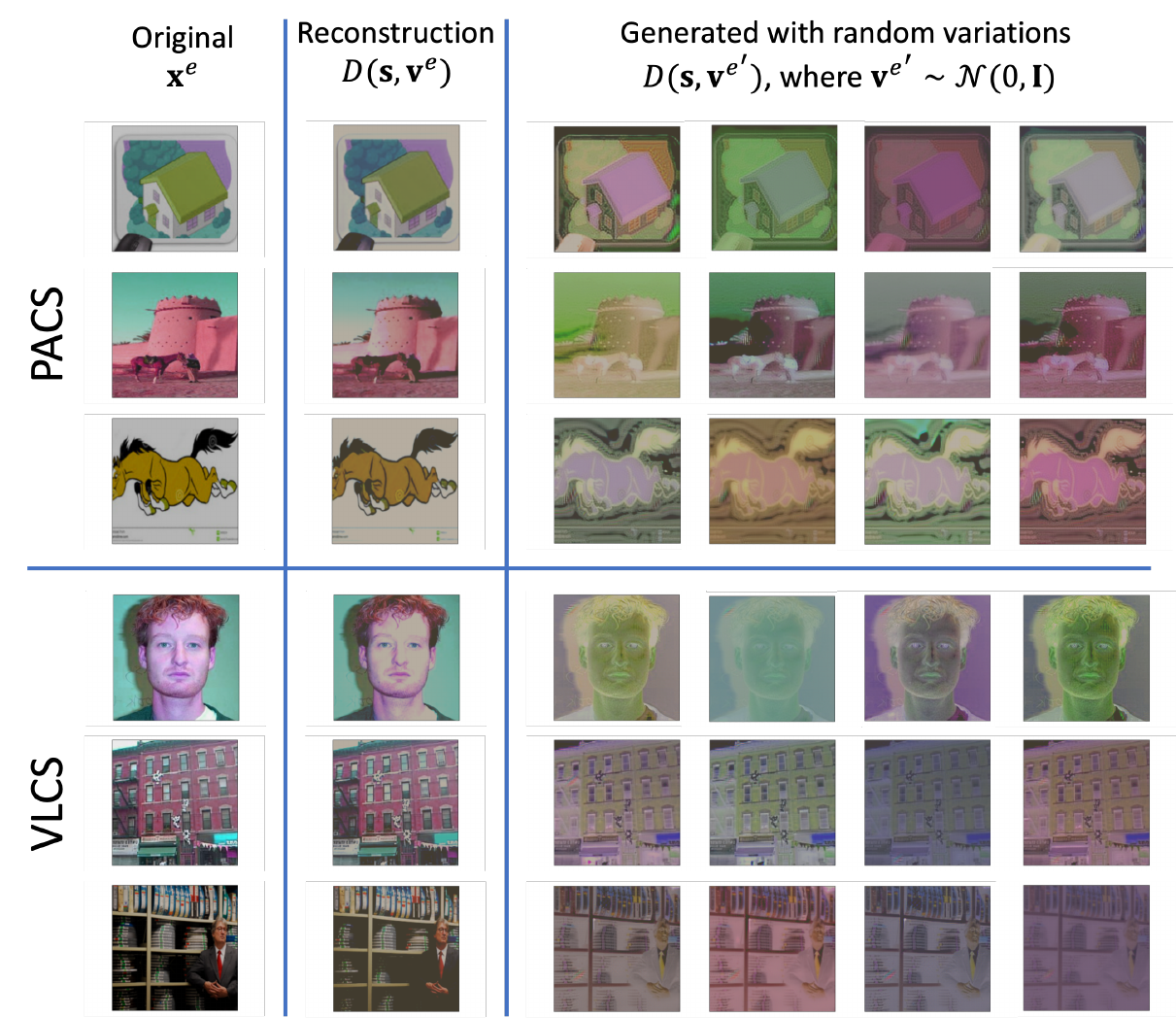}
    \caption{Reconstruction of the original images from latent factors and their variants by randomly changing variations through the pre-trained $G$ model.}
    \label{fig:styles}
\end{figure}

\textbf{Synthetic Domain Examples.} Through the transformation model $G$, we are able to transform data from a training domain to a randomly generated synthetic domain. This transformation is achieved by substituting the input's variation factor, $\mathbf{v}^e$, with a randomly sampled variation factor $\mathbf{v}^{e'}\sim \mathcal{N}(0,\mathbf{I})$. Figure \ref{fig:styles} illustrates how $G$ proficiently reconstructs $\mathbf{x}^e$ using its semantic factor $\mathbf{s}$ and variation factor $\mathbf{v}^e$. Moreover, $G$ is capable of transforming $\mathbf{x}^e$ into diverse synthetic domains characterized by random variations. These capabilities collectively contribute to a more robust $G$-invariance in semantic representations.

\begin{figure}[!t]
    \centering
    \includegraphics[width=0.8\linewidth]{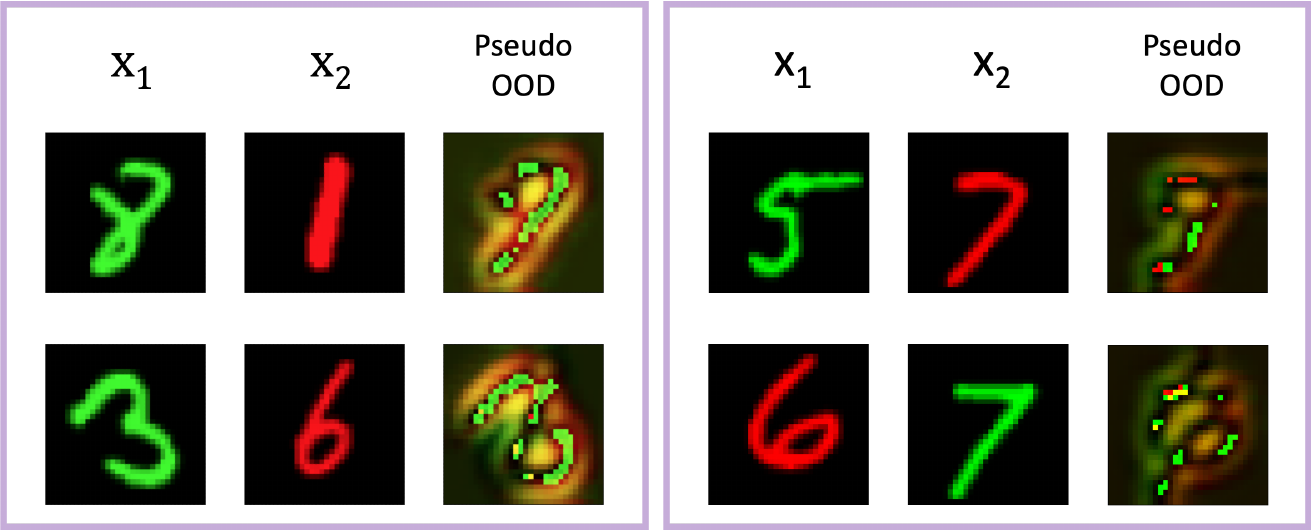}
    \caption{Examples of pseudo-OODs on the \textsc{ColoredMNIST} dataset, exhibiting densities below 0.1, with digit 9 designated as the OOD class.}
    \label{fig:oods}
\end{figure}

\textbf{Pseudo-OODs.} Figure \ref{fig:oods} showcases two pseudo-OODs with densities below 0.1 on the \textsc{ColoredMNIST} dataset, digit 9 is designated as the OOD class. As evident from the figure, these pseudo-OODs capture unobserved semantics distinct from the training data.

\textbf{OOD Detection Performance.} For OOD detection performance, we tabulated the mean AUROC values across trials, OOD class selections, and domains in Table \ref{tab:ood-summary}, corresponding to \textsc{ColoredMNIST}, \textsc{PACS}, and \textsc{VLCS}, respectively. The results indicate that \sysname{} consistently outperforms other methods. Notably, on the \textsc{ColoredMNIST} dataset, \sysname{} achieved an average AUROC of \num{77.83}, which is nearly 20\% higher than the next best baseline method. For the \textsc{PACS} and \textsc{VLCS} datasets, \sysname{}'s performance surpassed the best baseline method by margins of approximately 12\% to 13\%.

\begin{table}[!t]
\centering
\setlength\tabcolsep{5pt}
\caption{OOD Detection Performance Evaluation: All AUROC values represent the average across all trials, OOD class selections, and test domains. Top results are highlighted in \textbf{bold}.}
\label{tab:ood-summary}
\begin{tabular}{c|l|c|c|c|c|c}
\toprule
\multirow{3}{*}{Data} &  & \multicolumn{5}{c}{\textbf{OOD Detection, AUROC}}  \\
\cmidrule{3-7}
& & \multicolumn{5}{c}{OOD Detection Algorithms}  \\
\midrule
& Methods & MSP    & Energy & DDU & OCSVM & Avg \\ 
\cmidrule{2-7}
\parbox[t]{2mm}{\multirow{5}{*}{\rotatebox[origin=c]{90}{\textsc{\scriptsize CMNIST}}}} 
& ERM    & 48.93 $\pm$ 1.36 & 50.39 $\pm$ 0.08 & 52.17 $\pm$ 2.58 & 51.79 $\pm$ 0.69 & 50.82 \\
& IRM    & 51.11 $\pm$ 3.11 & 49.24 $\pm$ 1.47 & 50.16 $\pm$ 1.24 & 52.99 $\pm$ 0.24 & 50.87 \\
& Mixup  & 50.25 $\pm$ 1.42 & 49.64 $\pm$ 1.25 & 52.12 $\pm$ 0.61 & 53.23 $\pm$ 1.37 & 51.31 \\
& MBDG   & 58.46 $\pm$ 5.81 & 59.94 $\pm$ 5.69 & 62.90 $\pm$ 7.74 & 54.28 $\pm$ 1.40 & 58.90 \\
\cmidrule{2-7}
& Ours   & \textbf{81.81} $\pm$ 2.46 & \textbf{81.56} $\pm$ 2.55 & \textbf{74.90} $\pm$ 2.35 & \textbf{73.03} $\pm$ 1.96 & \textbf{77.83} \\
\bottomrule
 & Methods & MSP    & Energy & DDU & OCSVM & Avg \\
\cmidrule{2-7}
\parbox[t]{2mm}{\multirow{5}{*}{\rotatebox[origin=c]{90}{\textsc{\scriptsize PACS}}}} 
& ERM   & 78.04 $\pm$ 1.75 & 80.12 $\pm$ 1.87 & 55.59 $\pm$ 4.64 & 49.54 $\pm$ 1.76 & 65.82  \\ 
& IRM   & 78.05 $\pm$ 4.95 & 81.07 $\pm$ 4.48 & 50.48 $\pm$ 5.44 & 56.42 $\pm$ 3.74 & 66.50  \\ 
& Mixup & 74.70 $\pm$ 4.49 & 74.14 $\pm$ 4.34 & 69.02 $\pm$ 2.34 & 35.29 $\pm$ 2.35 & 63.29  \\ 
& MBDG  & 76.43 $\pm$ 3.26 & 78.51 $\pm$ 3.74 & 53.90 $\pm$ 7.82 & 56.50 $\pm$ 6.22 & 66.33  \\ 
\cmidrule{2-7}
& Ours  & \textbf{84.78} $\pm$ 2.90 & \textbf{87.89} $\pm$ 2.49 & \textbf{85.65} $\pm$ 2.78 & \textbf{58.70} $\pm$ 3.02 & \textbf{79.25}  \\ 
\bottomrule
 & Methods & MSP    & Energy & DDU & OCSVM & Avg \\
\cmidrule{2-7}
\parbox[t]{2mm}{\multirow{5}{*}{\rotatebox[origin=c]{90}{\textsc{\scriptsize VLCS}}}} 
& ERM    & 69.55 $\pm$ 8.25 & 72.04 $\pm$ 7.89 & 58.04 $\pm$ 9.34 & 35.46 $\pm$ 6.05 & 58.77  \\
& IRM    & 69.12 $\pm$ 7.57 & 72.54 $\pm$ 6.47 & 56.97 $\pm$ 9.39 & 35.19 $\pm$ 7.00 & 58.46  \\
& Mixup  & 65.20 $\pm$ 8.85 & 68.18 $\pm$ 7.63 & 59.87 $\pm$ 8.37 & 30.28 $\pm$ 7.09 & 55.88  \\
& MBDG   & 65.60 $\pm$ 7.10 & 71.89 $\pm$ 5.90 & 51.18 $\pm$ 4.31 & 62.78 $\pm$ 4.63 & 62.86  \\
\cmidrule{2-7}
& Ours   & \textbf{75.04} $\pm$ 7.35 & \textbf{78.64} $\pm$ 6.38 & \textbf{80.14} $\pm$ 6.57 & \textbf{68.25} $\pm$ 8.07 & \textbf{75.52}  \\
\bottomrule
\end{tabular}
\end{table}

\begin{table}[!t]
\centering
\setlength\tabcolsep{5pt}
\caption{Classification Performance Evaluation: All accuracy values presented are averaged over all trials and OOD class choices. Top results are highlighted in \textbf{bold}.}
\label{tab:classification-summary}
\begin{tabular}{c|l|c|c|c|c|c}
\toprule
\multirow{3}{*}{Data} &  & \multicolumn{4}{c}{\textbf{InD Classification, Accuracy}} \\
\cmidrule{3-7}
&  & \multicolumn{4}{c}{Test Domains} \\
\midrule
& Methods & +90\% & +80\% & -90\% & & Avg\\ 
\cmidrule{2-7}
\parbox[t]{2mm}{\multirow{5}{*}{\rotatebox[origin=c]{90}{\textsc{\scriptsize CMNIST}}}} 
& ERM    & 43.91 $\pm$ 3.72 & 49.69 $\pm$ 1.47 & 10.21 $\pm$ 0.26 & & 34.71\\
& IRM    & 51.63 $\pm$ 1.85 & 59.84 $\pm$ 1.55 & 12.85 $\pm$ 2.15 & & 41.52\\
& Mixup  & 50.02 $\pm$ 0.77 & 48.85 $\pm$ 1.15 & 10.17 $\pm$ 0.07 & & 36.44\\
& MBDG   & 68.67 $\pm$ 2.76 & 70.54 $\pm$ 2.91 & 19.16 $\pm$ 3.30 & & 52.87\\
\cmidrule{2-7}
& Ours   & \textbf{73.10} $\pm$ 0.37 & \textbf{72.51} $\pm$ 0.44 & \textbf{67.07} $\pm$ 3.17 & & \textbf{70.94}\\
\bottomrule
 & Methods & A & C & P & S & Avg \\
\cmidrule{2-7}
\parbox[t]{2mm}{\multirow{5}{*}{\rotatebox[origin=c]{90}{\textsc{\scriptsize PACS}}}} 
& ERM   & 87.92 $\pm$ 1.60 & 79.36 $\pm$ 1.67 & 96.31 $\pm$ 0.93 & 74.34 $\pm$ 0.37 & 84.53 \\ 
& IRM   & 86.90 $\pm$ 1.31 & 79.22 $\pm$ 1.79 & 95.16 $\pm$ 0.64 & 74.41 $\pm$ 3.07 & 83.99 \\ 
& Mixup & 88.93 $\pm$ 0.91 & 79.80 $\pm$ 1.55 & \textbf{96.82} $\pm$ 0.60 & 74.13 $\pm$ 2.52 & 84.98 \\ 
& MBDG  & 82.67 $\pm$ 0.85 & 72.87 $\pm$ 1.03 & 80.7 $\pm$ 10.87 & \textbf{81.34} $\pm$ 1.53 & 79.65 \\ 
\cmidrule{2-7}
& Ours  & \textbf{89.64} $\pm$ 1.44 & \textbf{81.31} $\pm$ 2.26 & 96.12 $\pm$ 0.43 & 76.20 $\pm$ 2.18 & \textbf{85.91} \\ 
\bottomrule
 & Methods & C & L & S & V & Avg \\
\cmidrule{2-7}
\parbox[t]{2mm}{\multirow{5}{*}{\rotatebox[origin=c]{90}{\textsc{\scriptsize VLCS}}}} 
& ERM    & 99.15 $\pm$ 0.01 & 64.43 $\pm$ 1.04 & 75.95 $\pm$ 0.40 & \textbf{81.04} $\pm$ 0.86 & \textbf{80.27} \\
& IRM    & 98.26 $\pm$ 0.21 & 64.57 $\pm$ 1.19 & 73.08 $\pm$ 3.43 & 78.20 $\pm$ 2.91 & 78.71 \\
& Mixup  & 98.83 $\pm$ 0.03 & \textbf{65.90} $\pm$ 1.40 & 75.81 $\pm$ 0.12 & 79.21 $\pm$ 2.84 & 79.95 \\
& MBDG   & \textbf{99.23} $\pm$ 0.12 & 64.10 $\pm$ 0.43 & 71.02 $\pm$ 0.70 & 77.02 $\pm$ 0.01 & 77.87 \\
\cmidrule{2-7}
& Ours   & 98.37 $\pm$ 0.73 & 62.90 $\pm$ 0.50 & \textbf{76.60} $\pm$ 1.10 & 75.00 $\pm$ 3.08 & 78.30 \\
\bottomrule
\end{tabular}
\end{table}

Among the three datasets, \textsc{ColoredMNIST} presents the greatest challenge. This complexity arises from a 25\% manually-inserted label error and a strong correlation between the color of the digit and the image label. Consequently, all baseline methods struggled to achieve satisfactory OOD detection performance.

\sysname{}, on the other hand, effectively navigates the mislabeling issues in the training set. It does so by leveraging the energy score disparities between the in-distribution (InD) training samples and the pseudo-OODs. Specifically, the semantic OOD detection regularization term, denoted as $R_{OOD}$, in \sysname{} mandates a distinct energy difference between InD and OOD inputs. By design, $R_{OOD}$ is indifferent to the specific labels; it solely discerns whether an input is InD or pseudo-OOD. Given that the mislabeled training samples retain their InD labels, \sysname{} demonstrates remarkable performance on \textsc{ColoredMNIST}. The nuances of this superiority are further elucidated in the ablation study.

\begin{table}[!t]
\centering
\footnotesize
\caption{Ablation Study Performance Evaluation: The reported AUROC values represent the average across all trials, OOD class selections, and test domains. The top-performing results are highlighted in \textbf{bold}.}
\label{tab:ablation}
\begin{tabular}{c|l|c|c|c|c|c}
\toprule
Data & Settings & MSP & Energy & DDU & OCSVM & Avg \\
\midrule
\parbox[t]{2mm}{\multirow{3}{*}{\rotatebox[origin=c]{90}{\textsc{\scriptsize CMNIST}}}} 
& \sysname{} w/o $R_{OOD}$\&$R_{DG}$           
& 48.93 $\pm$\scriptsize 1.36 & 50.39 $\pm$\scriptsize 0.08 & 52.17 $\pm$\scriptsize 2.58 & 51.79 $\pm$\scriptsize 0.69 & 50.82 \\
& \sysname{} w/o $R_{OOD}$      
& 67.38 $\pm$\scriptsize 1.55 & 67.29 $\pm$\scriptsize 1.28 & 73.74 $\pm$\scriptsize 2.81 & 59.61 $\pm$\scriptsize 4.59 & 67.01 \\
& \sysname{} (full) 
& \textbf{81.81} $\pm$\scriptsize 2.46 & \textbf{81.56} $\pm$\scriptsize 2.55 & \textbf{74.90} $\pm$\scriptsize 2.35 & \textbf{73.03} $\pm$\scriptsize 1.96 & \textbf{77.83} \\
\bottomrule
\parbox[t]{2mm}{\multirow{3}{*}{\rotatebox[origin=c]{90}{\textsc{\scriptsize PACS}}}} 
& \sysname{} w/o $R_{OOD}$\&$R_{DG}$ 
& 78.04 $\pm$\scriptsize 1.75 & 80.12 $\pm$\scriptsize 1.87 & 55.59 $\pm$\scriptsize 4.64 & 49.54 $\pm$\scriptsize 1.76 & 65.82 \\
& \sysname{} w/o $R_{OOD}$ 
& 78.84 $\pm$\scriptsize 2.90 & 80.91 $\pm$\scriptsize 3.10 & 72.99 $\pm$\scriptsize 2.60 & \textbf{65.60} $\pm$\scriptsize 2.81 & 74.59 \\
& \sysname{} (full) 
& \textbf{84.78} $\pm$\scriptsize 2.90 & \textbf{87.89} $\pm$\scriptsize 2.49 & \textbf{85.65} $\pm$\scriptsize 2.78 & 58.70 $\pm$\scriptsize 3.02 & \textbf{79.25} \\
\bottomrule
\parbox[t]{2mm}{\multirow{3}{*}{\rotatebox[origin=c]{90}{\textsc{\scriptsize VLCS}}}} 
& \sysname{} w/o $R_{OOD}$\&$R_{DG}$ 
& 69.55 $\pm$\scriptsize 8.25 & 72.04 $\pm$\scriptsize 7.89 & 58.04 $\pm$\scriptsize 9.34 & 35.46 $\pm$\scriptsize 6.05 & 58.77 \\
& \sysname{} w/o $R_{OOD}$ 
& 69.19 $\pm$\scriptsize 8.83 & 72.46 $\pm$\scriptsize 7.94 & 60.87 $\pm$\scriptsize 6.93 & \textbf{69.45} $\pm$\scriptsize 6.00 & 68.00 \\
& \sysname{}(full) 
& \textbf{75.04} $\pm$\scriptsize 7.35 & \textbf{78.64} $\pm$\scriptsize 6.38 & \textbf{80.14} $\pm$\scriptsize 6.57 & 68.25 $\pm$\scriptsize 8.07 & \textbf{75.52} \\
\bottomrule
\end{tabular}
\end{table}



\textbf{InD Classification Performance.} Table \ref{tab:classification-summary} presents the mean classification accuracy across trials and OOD class selections along with the standard error. Despite introducing domain regularization and semantic OOD detection regularization, \sysname{} not only maintained but in some cases even enhanced its InD classification performance across all three datasets.

For \textsc{ColoredMNIST}, \sysname{}'s classification accuracy on test domains +90\% and +80\% was on par with the top-performing baseline, MBDG. However, in the test domain -90\%, \sysname{} outperformed other baselines with notably higher classification accuracy.

The challenge for most baselines in domain -90\% stems from its use as a test domain. When domains +90\% and +80\% serve as training domains, both domains exhibit a strong positive correlation between the digit color and label. This complicates the task of training a neural network that can generalize effectively to a test domain exhibiting the opposite correlation.

Yet, \sysname{} manages this feat successfully by ensuring semantic invariance via domain regularization, making it more resilient against such color-label correlations. MBDG shares a similar objective of achieving semantic invariance by utilizing data from random domains. However, it penalizes semantic domain variance in a low-dimensional output space, namely the class probabilities. This approach restricts its representational capacity. In contrast, \sysname{} applies regularization in a high-dimensional feature space, capturing more intricate information. The enhancement in classification accuracy, as evident in Table \ref{tab:classification-summary}, underscores the advantage of using a high-dimensional feature space for learning semantic invariance, especially in domains with color-label correlations.

On both \textsc{PACS} and \textsc{VLCS} datasets, \sysname{} consistently matched or outperformed the classification accuracy of other methods for each test domain.

\textbf{Ablation Studies.} To elucidate both the individual and cumulative effects of the components within our \sysname{} framework, we embarked on a systematic ablation study using three different settings.


The results from Table \ref{tab:ablation} reveal that without $R_{OOD}$ and $R_{DG}$, \sysname{} produced subpar outcomes on all three datasets. This underperformance was particularly evident for the \textsc{ColoredMNIST} dataset and when using DDU and OCSVM.
By using $R_{DG}$ only (\sysname{} w/o $R_{OOD}$), it leads to significant enhancements in semantic OOD detection. The average AUROC increased by 16.19\% on \textsc{ColoredMNIST}, 8.77\% on \textsc{PACS}, and 9.23\% \textsc{VLCS}.
Lastly, compared with the previous setting \sysname{} w/o $R_{OOD}$, the average AUROC in the full model of \sysname{} increases by 10.82\% on \textsc{ColoredMNIST}, 4.66\% on \textsc{PACS}, and 7.52\% on \textsc{VLCS}.

\textbf{Feature Space Regularization versus Output Space Regularization.}
In this paper, the semantic $G$-invariance is learned through feature space regularization. Specifically, we employ the domain generalization regularization term $R_{DG}$ to enforce the semantic between $(\mathbf{x}^{e},y)$ and the random domain counterpart $(\mathbf{x}^{e'},y)$ in the feature space. One might wonder why we prioritize the feature space over the output space for learning semantic $G$-invariance. The primary reason is that the feature space representations contain richer information than the final outputs. As highlighted by recent neural network interpretation studies \cite{zhou2018interpreting,rauker2023toward}, deeper layers in a neural network tend to capture increasingly complex concepts. As for the output space, constrained by its feature dimension, typically represents high-level conceptual semantics, thereby limiting its representational capacity. Furthermore, OOD can take many forms, each with unique characteristics. Some OODs might exhibit only subtle deviations from the in-distribution data. To identify such nuanced differences, it becomes imperative to leverage features with higher representational capacity.

\begin{figure}[!t]
\centering
    \begin{subfigure}[b]{0.48\textwidth}
       \includegraphics[width=1\linewidth]{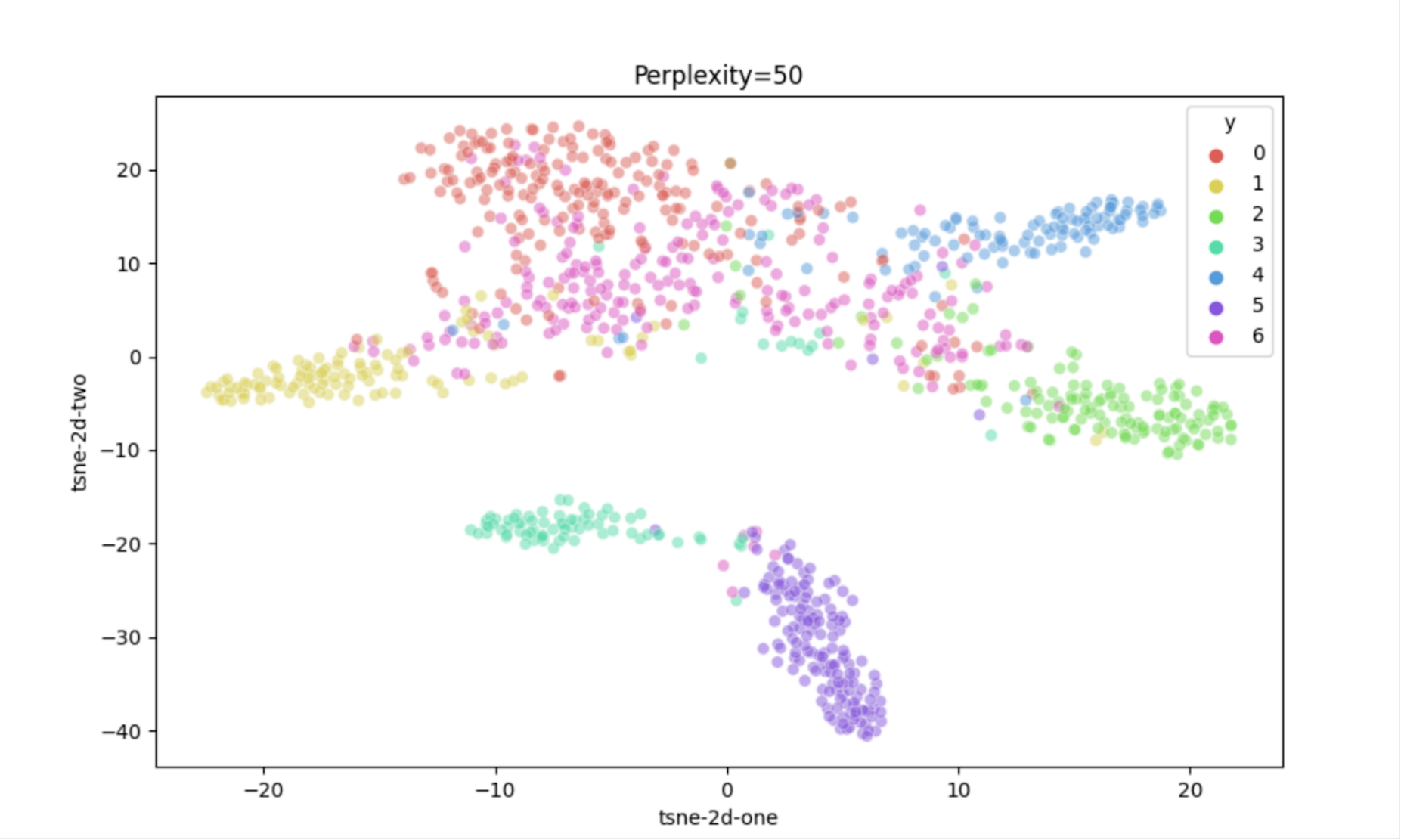}
       \caption{Output space regularization}
       \label{fig:Ng1} 
    \end{subfigure}
    \begin{subfigure}[b]{0.48\textwidth}
       \includegraphics[width=1\linewidth]{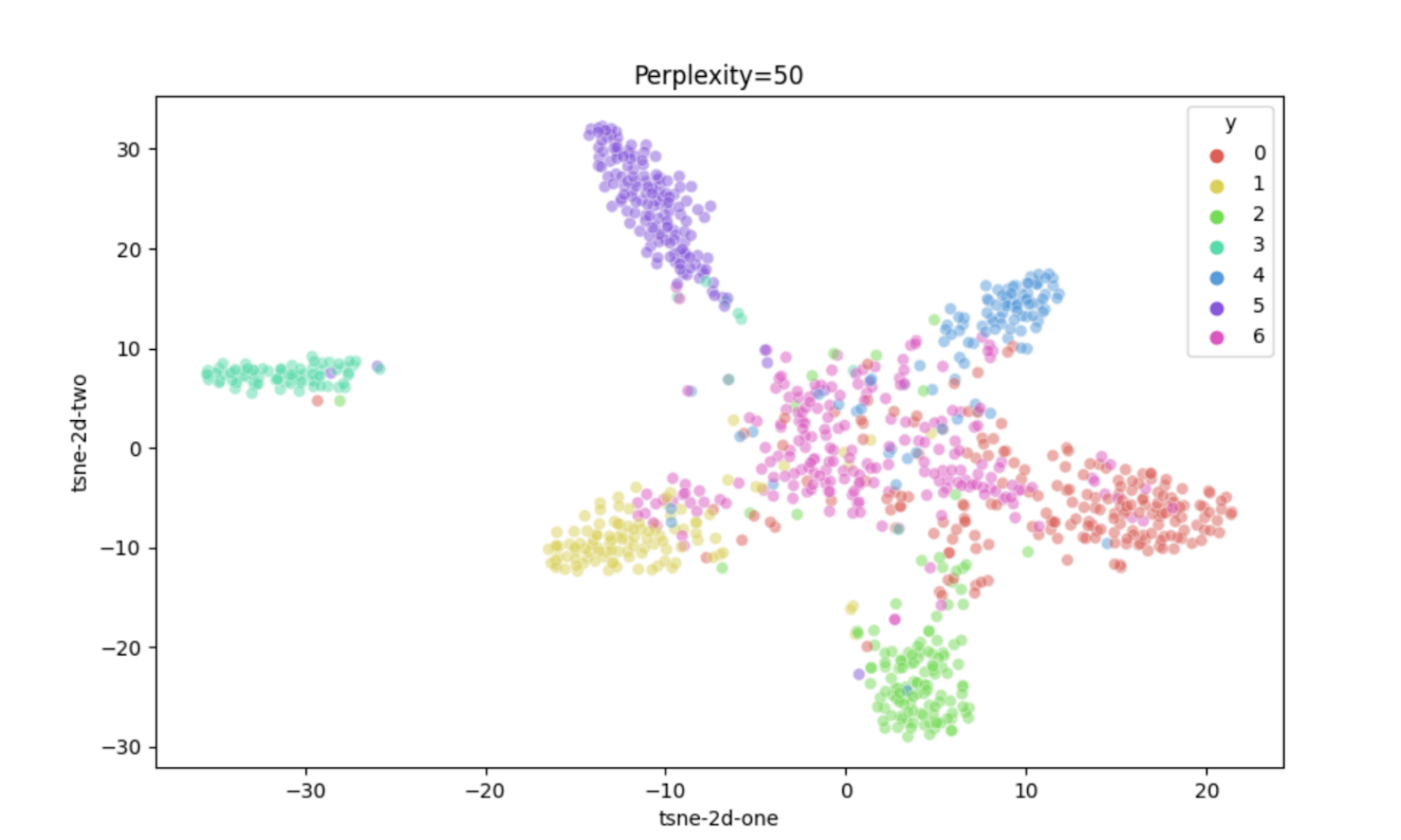}
       \caption{Feature space regularization}
       \label{fig:Ng2}
    \end{subfigure}
    
    \caption{Comparison of t-SNE plots for latent features of models using output space regularization versus feature space regularization. These plots are generated on the \textsc{PACS} dataset, with "Art Painting" as the test domain. Class 6 (pink) is the designated OOD class.}
\label{fig:reg-compare}
\end{figure}

To provide a comparative perspective, we compare \sysname{}, which employs feature space regularization, against a model where the $R_{DG}$ term is substituted with output space regularization. Their respective t-SNE visualizations, derived from the PACS dataset with "Art Painting" as the test domain and class 6 as OODs, are presented in Figure \ref{fig:reg-compare}. It is evident that the feature space regularization approach provides clearer boundaries for each class cluster. Moreover, the OOD samples (pink points) mostly lie in the middle with less overlapping with InD clusters.

\textbf{OOD Detection Performance Details.} In Tables \ref{tab:ood-details-1}, \ref{tab:ood-details-2}, and \ref{tab:ood-details-3}, we report the detailed AUROC values for each test domain. From these results, it is evident that \sysname{} consistently stands out, delivering superior performance across the majority of configurations.

\begin{table}[!t]
\centering
\caption{\textsc{ColoredMNIST} Performance Evaluation Details: The AUROC values are the average across all trials, OOD class selections. Top results are highlighted in \textbf{bold}.}
\begin{tabular}{c|l|c|c|c|c}
\toprule
\multirow{3}{*}{Data} &  & \multicolumn{4}{c}{\textbf{OOD Detection, AUROC}} \\
\cmidrule{3-6}
& & \multicolumn{4}{c}{OOD Detection Algorithms}  \\
\cmidrule{1-6}
& Methods & MSP    & Energy & DDU & OCSVM  \\ 
\cmidrule{2-6}
\parbox[t]{2mm}{\multirow{5}{*}{\rotatebox[origin=c]{90}{+90\%}}} 
& ERM    & 49.72 $\pm$ 2.29 & 50.54 $\pm$ 2.74 & 50.54 $\pm$ 5.63 & 53.17 $\pm$ 2.80 \\
& IRM    & 57.20 $\pm$ 2.21 & 51.94 $\pm$ 6.06 & 52.36 $\pm$ 2.05 & 53.17 $\pm$ 0.49 \\
& Mixup  & 53.00 $\pm$ 1.74 & 49.50 $\pm$ 2.45 & 53.20 $\pm$ 4.43 & 55.84 $\pm$ 1.90 \\
& MBDG   & 62.94 $\pm$ 3.60 & 66.34 $\pm$ 4.92 & 71.54 $\pm$ 8.99 & 56.72 $\pm$ 4.34 \\
\cmidrule{2-6}
& Ours   & \textbf{86.19} $\pm$ 2.13 & \textbf{86.28} $\pm$ 2.09 & \textbf{79.61} $\pm$ 3.24 & \textbf{76.95} $\pm$ 2.86  \\
\midrule
\parbox[t]{2mm}{\multirow{5}{*}{\rotatebox[origin=c]{90}{+80\%}}} 
& ERM    & 50.78 $\pm$ 1.72 & 50.38 $\pm$ 3.49 & 48.74 $\pm$ 5.79 & 51.10 $\pm$ 2.17  \\
& IRM    & 49.16 $\pm$ 1.57 & 48.87 $\pm$ 3.66 & 48.06 $\pm$ 3.76 & 52.51 $\pm$ 3.03  \\
& Mixup  & 49.50 $\pm$ 2.08 & 51.88 $\pm$ 2.27 & 52.08 $\pm$ 3.44 & 52.65 $\pm$ 3.09  \\
& MBDG   & 65.52 $\pm$ 5.91 & 64.89 $\pm$ 5.61 & 69.70 $\pm$ 4.15 & 54.26 $\pm$ 4.25  \\
\cmidrule{2-6}
& Ours   & \textbf{81.58} $\pm$ 2.67 & \textbf{80.88} $\pm$ 2.46 & \textbf{72.48} $\pm$ 5.30 & \textbf{70.89} $\pm$ 3.08  \\
\midrule
\parbox[t]{2mm}{\multirow{5}{*}{\rotatebox[origin=c]{90}{-90\%}}} 
& ERM    & 46.28 $\pm$ 2.09 & 50.26 $\pm$ 3.60 & 57.23 $\pm$ 5.34 & 51.11 $\pm$ 2.19  \\
& IRM    & 46.98 $\pm$ 1.55 & 46.90 $\pm$ 2.39 & 50.05 $\pm$ 3.54 & 53.28 $\pm$ 1.21  \\
& Mixup  & 48.26 $\pm$ 1.17 & 47.55 $\pm$ 3.22 & 51.09 $\pm$ 2.95 & 51.20 $\pm$ 3.66  \\
& MBDG   & 46.93 $\pm$ 4.22 & 48.59 $\pm$ 3.11 & 47.46 $\pm$ 6.35 & 51.86 $\pm$ 3.98  \\
\cmidrule{2-6}
& Ours   & \textbf{77.67} $\pm$ 1.97 & \textbf{77.53} $\pm$ 1.80 & \textbf{72.62} $\pm$ 3.66 & \textbf{71.24} $\pm$ 4.96  \\
\bottomrule
\end{tabular}
\label{tab:ood-details-1}
\end{table}

\begin{table}[!t]
\centering
\caption{\textsc{PACS} Performance Evaluation Details: The AUROC values are the average across all trials, OOD class selections. Top results are highlighted in \textbf{bold}.}
\begin{tabular}{c|l|c|c|c|c}
\toprule
\multirow{3}{*}{Data} &  & \multicolumn{4}{c}{\textbf{OOD Detection, AUROC}} \\
\cmidrule{3-6}
& & \multicolumn{4}{c}{OOD Detection Algorithms}  \\
\midrule
& Methods & MSP    & Energy & DDU & OCSVM  \\ 
\cmidrule{2-6}
\parbox[t]{2mm}{\multirow{5}{*}{\rotatebox[origin=c]{90}{Art Painting}}} 
& ERM    & 79.79 $\pm$ 4.28 & 80.00 $\pm$ 5.81 & 53.70 $\pm$ 7.56 & 50.15 $\pm$ 5.27  \\
& IRM    & 77.08 $\pm$ 5.31 & 78.05 $\pm$ 7.22 & 48.38 $\pm$ 12.87 & 58.81 $\pm$ 4.86  \\
& Mixup  & 75.17 $\pm$ 5.46 & 75.14 $\pm$ 6.07 & 70.06 $\pm$ 3.98 & 38.53 $\pm$ 3.25  \\
& MBDG   & 75.70 $\pm$ 1.86 & 75.62 $\pm$ 4.02 & 48.35 $\pm$ 10.66 & \textbf{61.09} $\pm$ 6.14  \\
\cmidrule{2-6}
& Ours   & \textbf{83.02} $\pm$ 3.52 & \textbf{84.50} $\pm$ 4.95 & \textbf{81.15} $\pm$ 4.67 & 52.47 $\pm$ 7.69  \\
\midrule
\parbox[t]{2mm}{\multirow{5}{*}{\rotatebox[origin=c]{90}{Cartoon}}} 
& ERM    & 73.83 $\pm$ 2.55 & 74.90 $\pm$ 4.06 & 65.79 $\pm$ 14.52 & 44.38 $\pm$ 16.92  \\
& IRM    & 69.06 $\pm$ 8.36 & 72.18 $\pm$ 9.50 & 64.42 $\pm$ 7.50 & 45.96 $\pm$ 12.59  \\
& Mixup  & 66.83 $\pm$ 4.79 & 64.28 $\pm$ 4.25 & 75.01 $\pm$ 5.61 & 29.09 $\pm$ 7.13  \\
& MBDG   & 68.72 $\pm$ 4.69 & 69.28 $\pm$ 5.37 & 63.72 $\pm$ 0.68 & 46.63 $\pm$ 4.91  \\
\cmidrule{2-6}
& Ours   & \textbf{80.32} $\pm$ 2.57 & \textbf{84.16} $\pm$ 2.59 & \textbf{83.59} $\pm$ 4.45 & \textbf{64.85} $\pm$ 9.77  \\
\midrule
\parbox[t]{2mm}{\multirow{5}{*}{\rotatebox[origin=c]{90}{Photo}}} 
& ERM    & 81.83 $\pm$ 7.00 & 83.22 $\pm$ 6.74 & 59.06 $\pm$ 11.86 & 51.73 $\pm$ 4.91  \\
& IRM    & 92.05 $\pm$ 0.38 & 93.41 $\pm$ 0.79 & 51.14 $\pm$ 12.63 & \textbf{63.61} $\pm$ 9.24  \\
& Mixup  & 87.13 $\pm$ 1.69 & 85.21 $\pm$ 2.78 & 64.13 $\pm$ 22.94 & 39.33 $\pm$ 24.24  \\
& MBDG   & 84.67 $\pm$ 4.76 & 85.37 $\pm$ 5.06 & 69.04 $\pm$ 7.78 & 46.25 $\pm$ 2.10  \\
\cmidrule{2-6}
& Ours   & \textbf{93.31} $\pm$ 2.78 & \textbf{94.87} $\pm$ 2.82 & \textbf{93.75} $\pm$ 2.99 & 54.68 $\pm$ 6.24  \\
\midrule
\parbox[t]{2mm}{\multirow{5}{*}{\rotatebox[origin=c]{90}{Sketch}}} 
& ERM    & 76.71 $\pm$ 4.48 & 82.38 $\pm$ 5.30 & 43.82 $\pm$ 1.94 & 51.89 $\pm$ 4.83  \\
& IRM    & 74.02 $\pm$ 0.59 & 80.64 $\pm$ 1.95 & 37.98 $\pm$ 8.25 & 57.32 $\pm$ 1.12  \\
& Mixup  & 69.68 $\pm$ 2.70 & 71.94 $\pm$ 0.63 & 66.88 $\pm$ 6.33 & 34.22 $\pm$ 10.10  \\
& MBDG   & 76.64 $\pm$ 2.01 & 83.76 $\pm$ 1.87 & 34.47 $\pm$ 8.42 & \textbf{72.03} $\pm$ 9.26  \\
\cmidrule{2-6}
& Ours   & \textbf{82.45} $\pm$ 1.87 & \textbf{88.02} $\pm$ 2.56 & \textbf{84.09} $\pm$ 3.71 & 62.82 $\pm$ 12.30  \\
\bottomrule
\end{tabular}
\label{tab:ood-details-2}
\end{table}

\begin{table}[!t]
\centering
\caption{\textsc{VLCS} Performance Evaluation Details: The AUROC values are the average across all trials, OOD class selections. Top results are highlighted in \textbf{bold}.}
\begin{tabular}{c|l|c|c|c|c}
\toprule
\multirow{3}{*}{Data} &  & \multicolumn{4}{c}{\textbf{OOD Detection, AUROC}} \\
\cmidrule{3-6}
& & \multicolumn{4}{c}{OOD Detection Algorithms}  \\
\cmidrule{1-6}
& Methods & MSP    & Energy & DDU & OCSVM  \\ 
\cmidrule{2-6}
\parbox[t]{2mm}{\multirow{5}{*}{\rotatebox[origin=c]{90}{Caltech101}}} 
& ERM    & 92.01 $\pm$ 4.20 & 93.62 $\pm$ 3.00 & 82.13 $\pm$ 9.01 & 19.34 $\pm$ 12.10  \\
& IRM    & 89.45 $\pm$ 7.11 & 90.44 $\pm$ 6.40 & 80.53 $\pm$ 7.60 & 16.26 $\pm$ 9.22  \\
& Mixup  & 89.60 $\pm$ 1.34 & 90.02 $\pm$ 0.37 & 81.26 $\pm$ 7.79 & 14.62 $\pm$ 0.56  \\
& MBDG   & 83.00 $\pm$ 6.66 & 87.46 $\pm$ 5.02 & 61.02 $\pm$ 3.56 & 75.58 $\pm$ 1.30  \\
\cmidrule{2-6}
& Ours   & \textbf{93.34} $\pm$ 4.39 & \textbf{94.54} $\pm$ 2.88 & \textbf{95.62} $\pm$ 2.63 & \textbf{90.00} $\pm$ 3.83  \\
\midrule
\parbox[t]{2mm}{\multirow{5}{*}{\rotatebox[origin=c]{90}{LabelMe}}} 
& ERM    & 71.49 $\pm$ 3.75 & 73.24 $\pm$ 2.62 & 37.30 $\pm$ 8.34 & 46.96 $\pm$ 17.42  \\
& IRM    & 71.82 $\pm$ 3.68 & 73.63 $\pm$ 4.31 & 37.68 $\pm$ 5.26 & 45.38 $\pm$ 9.26  \\
& Mixup  & 66.82 $\pm$ 10.78 & 67.34 $\pm$ 12.70 & 52.88 $\pm$ 20.02 & 28.06 $\pm$ 14.97  \\
& MBDG   & 71.17 $\pm$ 5.37 & 74.20 $\pm$ 4.82 & 41.32 $\pm$ 8.79 & 58.48 $\pm$ 14.13  \\
\cmidrule{2-6}
& Ours   & \textbf{80.52} $\pm$ 8.14 & \textbf{83.28} $\pm$ 10.02 & \textbf{84.30} $\pm$ 9.95 & \textbf{61.20} $\pm$ 4.89  \\
\midrule
\parbox[t]{2mm}{\multirow{5}{*}{\rotatebox[origin=c]{90}{Sun09}}} 
& ERM    & 54.97 $\pm$ 1.75 & 57.48 $\pm$ 1.95 & 60.26 $\pm$ 1.68 & 33.54 $\pm$ 2.38  \\
& IRM    & 56.53 $\pm$ 3.57 & 62.58 $\pm$ 0.62 & 62.62 $\pm$ 6.77 & 32.98 $\pm$ 8.14  \\
& Mixup  & 53.60 $\pm$ 6.39 & 57.02 $\pm$ 3.86 & 63.42 $\pm$ 6.52 & 29.39 $\pm$ 6.23  \\
& MBDG   & 56.04 $\pm$ 11.12 & 65.11 $\pm$ 6.70 & 54.96 $\pm$ 3.64 & 62.94 $\pm$ 1.90  \\
\cmidrule{2-6}
& Ours   & \textbf{62.98} $\pm$ 1.61 & \textbf{69.56} $\pm$ 1.90 & \textbf{76.12} $\pm$ 5.20 & \textbf{69.59} $\pm$ 4.00  \\
\midrule
\parbox[t]{2mm}{\multirow{5}{*}{\rotatebox[origin=c]{90}{VOC2007}}} 
& ERM    & 59.72 $\pm$ 8.13 & 63.81 $\pm$ 3.74 & 52.46 $\pm$ 22.50 & 42.00 $\pm$ 23.96  \\
& IRM    & 58.68 $\pm$ 7.69 & 63.52 $\pm$ 8.60 & 47.04 $\pm$ 22.12 & 46.15 $\pm$ 23.57  \\
& Mixup  & 50.78 $\pm$ 1.10 & 58.34 $\pm$ 1.19 & 41.92 $\pm$ 22.15 & 49.03 $\pm$ 26.46  \\
& MBDG   & 52.18 $\pm$ 3.28 & 60.78 $\pm$ 5.91 & 47.40 $\pm$ 4.91 & \textbf{54.12} $\pm$ 26.16  \\
\cmidrule{2-6}
& Ours   & \textbf{63.33} $\pm$ 2.29 & \textbf{67.16} $\pm$ 1.57 & \textbf{64.50} $\pm$ 16.30 & 52.22 $\pm$ 15.13  \\
\bottomrule
\end{tabular}
\label{tab:ood-details-3}
\end{table}


\section{Conclusion}
    Addressing domain shifts and semantic OOD detection simultaneously poses a significant challenge: identifying OODs within unseen test domains. To tackle this, our \sysname{} framework introduces two regularization terms. The domain generalization regularization enforces semantic invariance, enabling the framework to generalize across unseen test domains, and the semantic OOD detection regularization that boosts the framework's ability to discern unseen OODs. Comprehensive evaluations on datasets \textsc{ColoredMNIST}, \textsc{PACS}, and \textsc{VLCS} verified the effectiveness of \sysname{}. Notably, it consistently surpasses existing benchmarks in OOD detection performance while retaining comparable InD classification accuracy. This represents a notable stride forward in addressing the problem of semantic OOD detection across domains.

\bibliographystyle{ACM-Reference-Format}
\bibliography{reference}

\end{document}